%% file: main.tex
\documentclass[10pt,twocolumn,letterpaper]{article}

\usepackage{cvpr}              


\definecolor{cvprblue}{rgb}{0.21,0.49,0.74}
\usepackage[pagebackref,breaklinks,colorlinks,allcolors=cvprblue]{hyperref}

\usepackage{indentfirst}

\def\paperID{*****} 
\def\confName{CVPR}
\def\confYear{2025}

\title{Animate Anyone 2: High-Fidelity Character Image Animation with Environment Affordance}

\author{Li Hu$^*$ \quad Guangyuan Wang$^*$ \quad Zhen Shen \quad Xin Gao \quad Dechao Meng \quad Lian Zhuo \\
Peng Zhang \quad Bang Zhang \quad Liefeng Bo\\
Tongyi Lab, Alibaba Group\\
\small \url{https://humanaigc.github.io/animate-anyone-2/}
}

\begin{document}

\twocolumn[{%
\renewcommand\twocolumn[1][]{#1}%
\maketitle
\begin{center}
    \centering
    \captionsetup{type=figure}
    \includegraphics[width=1.0\textwidth]{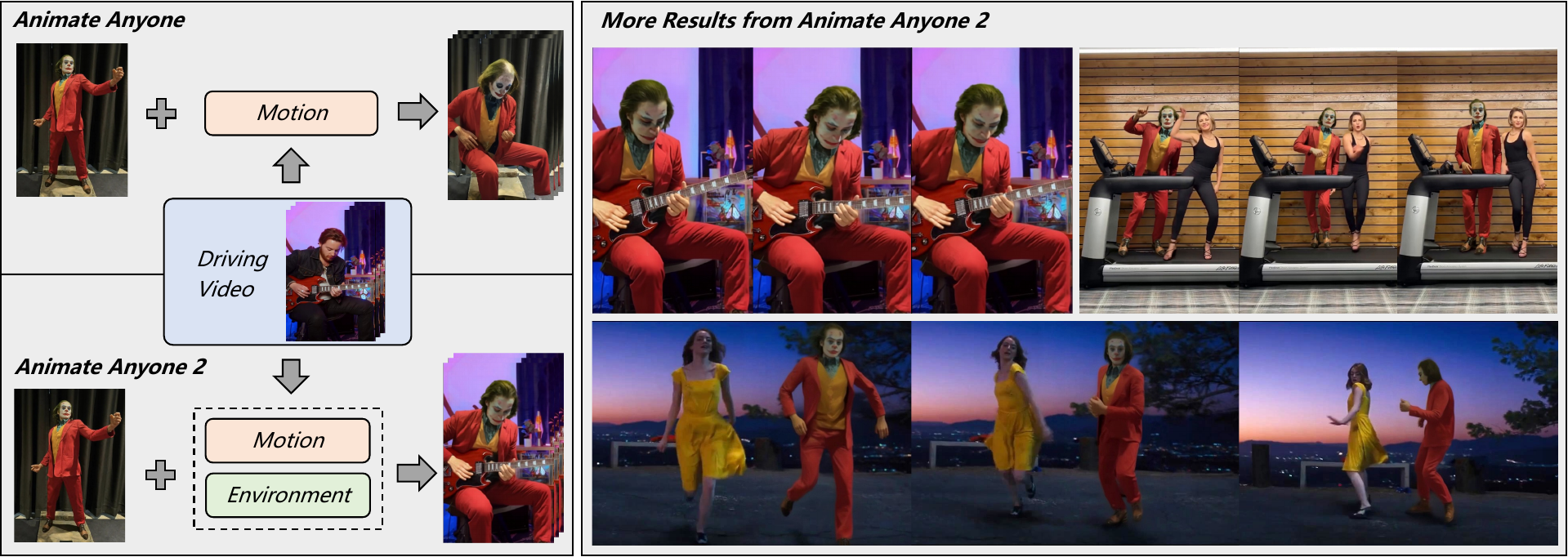}
    \captionof{figure}{We propose \textit{Animate Anyone 2}, which differs from previous character image animation methods that solely utilize motion signals to animate characters. Our approach additionally extracts environmental representations from the driving video, thereby enabling character animation to exhibit environment affordance. The generated results demonstrate that, beyond maintaining character consistency, \textit{Animate Anyone 2} can produce high-fidelity results that seamlessly integrate characters with the surrounding environment.}
    \label{fig:f1}
\end{center}%
}]

\maketitle

\begin{abstract}
Recent character image animation methods based on diffusion models, such as Animate Anyone, have made significant progress in generating consistent and generalizable character animations. However, these approaches fail to produce reasonable associations between characters and their environments. To address this limitation, we introduce Animate Anyone 2, aiming to animate characters with environment affordance. Beyond extracting motion signals from source video, we additionally capture environmental representations as conditional inputs. The environment is formulated as the region with the exclusion of characters and our model generates characters to populate these regions while maintaining coherence with the environmental context. We propose a shape-agnostic mask strategy that more effectively characterizes the relationship between character and environment. Furthermore, to enhance the fidelity of object interactions, we leverage an object guider to extract features of interacting objects and employ spatial blending for feature injection. We also introduce a pose modulation strategy that enables the model to handle more diverse motion patterns. Experimental results demonstrate the superior performance of the proposed method.
\end{abstract}

\renewcommand{\thefootnote}{}
\footnotetext{$^*$Equal contribution}

\section{Introduction}

The objective of character image animation is to synthesize animated video sequences utilizing a reference character image and a sequence of motion signals. 
Recent developments predominantly adopt diffusion-based frameworks~\cite{dreampose,disco,magicanimate,aa,magicpose,champ,unianimate,mimicmotion}, achieving notable enhancements in appearance consistency, motion stability and character generalizability. 
These advancements exhibit substantial potential in areas such as filmmaking, advertising, and virtual character applications.

In recent cross-identity animation workflows, motion signals are typically extracted from disparate videos, while the character's contextual environments are derived from static images. This setting introduces critical limitations: the spatial relationships between animated characters and their environments often lack authenticity, and intrinsic human-object interactions are disrupted. Consequently, most existing methods are predominantly limited to animating simple actions (e.g., individual gestures or dances) without adequately capturing the complex spatial and interactive relationships between characters and their surroundings. These limitations significantly hinder the advancement of character animation techniques.

Recent attempts to integrate character animation with scenes and objects, while promising, face significant challenges in generation quality and adaptability. 
For instance, MovieCharacter\cite{moviecharacter} synthesizes character videos by cascading the outputs from multiple algorithms, which introduces noticeable artifacts and unnatural visual discontinuities. 
AnchorCrafter\cite{anchorcrafter} primarily focuses on human-object manipulation animation, with relatively simplistic character motion and object appearance. 
MIMO\cite{mimo} addresses this challenge by composing characters, pre-processed backgrounds and occlusions, which are disentangled via depth. Such formulation for defining the relationship between characters and environments is suboptimal, 
limiting the ability to handle complex interactions.

In this paper, 
we propose to expand the scope of character animation by introducing \textit{Character Image Animation with Environment Affordance}.
Specifically, we define the research problem as follows: given a character image and a source video, the generated character animation should: 1) inherit character motion desired by the source video. 2) accurately demonstrate character-environment relationship consistent with the source video. 
This setting introduces novel challenges for character animation, as it requires that the model should effectively handle diverse and complex character motions, while ensuring precise interaction between characters and their environments throughout the animation process.

To achieve this, we introduce a novel framework \textit{Animate Anyone 2}. 
As illustrated in Fig.\ref{fig:f1}, unlike previous character animation methods that solely utilize motion signals, we additionally capture environmental representations from the source video as conditional inputs, which enables the model to learn the intrinsic relationship between character and environment in an end-to-end manner.
We formulate the environment by removing the character regions and our model generates characters to populate these regions while maintaining coherence with the environmental context. We develop a shape-agnostic mask strategy that better represents the boundary relationship between character and their contextual scenes, enabling effective learning for character-context integration while mitigating shape leakage issues. 
Second, to enhance the fidelity of object interactions, we introduce additional processing for interactive object regions. We design a lightweight object guider to extract interactive object features and propose a spatial blending mechanism to inject these features into the generation process. It facilitates the preservation of intricate interaction dynamics in the source video. 
Lastly, we propose depth-wise pose modulation approach for character motion modeling, empowering the model to handle more diverse and complex character poses with enhanced robustness.

The results in Fig.\ref{fig:f1} exhibit both high-quality character animation performance and remarkable environment affordance, manifested through three key advantages: 1) seamless scene integration; 2) coherent object interaction; and 3) robust handling of diverse and complex motions. Our approach is evaluated on corresponding benchmarks, achieving superior character animation results compared to existing methods. 
In summary, we highlight three key contributions of our paper.
\begin{itemize}
\item We introduce \textit{Animate Anyone 2}, a framework capable of animating character with environment affordance, achieving robust performance.
\item We propose a novel environment formulation and object injection strategy to achieve seamless character-environment integration.
\item We propose pose modulation strategy to enhance model robustness in challenging action scenarios.
\end{itemize}

\begin{figure*}[!t]
\begin{center}
	\setlength{\fboxrule}{0pt}
	\fbox{\includegraphics[width=0.99\textwidth]{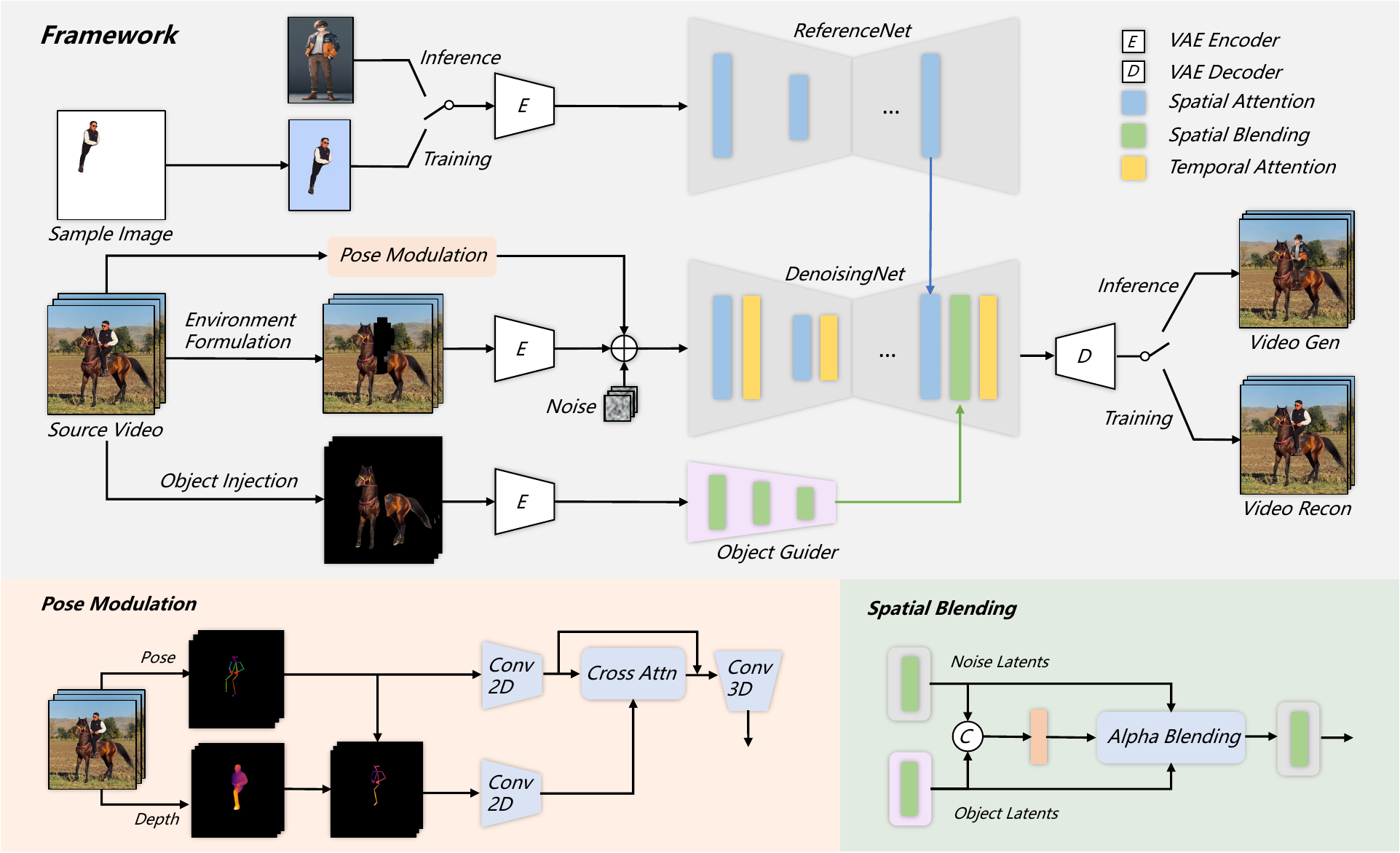}}
\end{center}
\vspace{-0.6cm}
\caption{The framework of \textit{Animate Anyone 2}. We capture environmental information from the source video. The environment is formulated as regions devoid of characters and incorporated as model input, enabling end-to-end learning of character-environment fusion. To preserve object interactions, we additionally inject features of objects interacting with the character. These object features are extracted by a lightweight object guider and merged into the denoising process via spatial blending. To handle more diverse motions, we propose a pose modulation approach to better represent the spatial relationships between body limbs. 
}
\vspace{-0.2cm}
\label{fig:overview}
\end{figure*}

\section{Related Works}

\subsection{Character Image Animation}
Distinguished from GAN-based\cite{gan,wgan,stylegan} approaches\cite{fomm,mraa,ren2020deep,tpsmm,siarohin2019animating,zhang2022exploring,bidirectionally,everybody}, diffusion-based image animation methods\cite{dreampose,disco,aa,magicanimate,magicpose,mimicmotion,champ,unianimate,tcan,tpc} have emerged as the current research mainstream. As the most representative approach, Animate Anyone\cite{aa} designs its framework based on Stable Diffusion\cite{ldm}, and the denoising network is structured as a 3D UNet\cite{align,animatediff} for temporal modeling. It proposes ReferenceNet, a symmetric UNet\cite{unet} architecture, to preserve appearance consistency and employs pose guider to incorporate skeleton information as driving signals for stable motion control. The Animate Anyone framework achieves robust and generalizable character animation, from which we extensively drew inspiration.

Some works propose improvements upon foundational frameworks. MimicMotion\cite{mimicmotion} leverages pretrained image-to-video capabilities of Stable Video Diffusion\cite{svd}, designing a PoseNet to inject skeleton information. UniAnimate\cite{unianimate} stacks reference images across temporal dimensions, utilizing mamba-based\cite{mamba} temporal modeling techniques. Some works explore different motion control signals. DisCo\cite{disco} and MagicAnimate\cite{magicanimate} utilizes DensePose\cite{densepose} as human body representations. Champ\cite{champ} employs the 3D parametric human model SMPL\cite{smpl}, integrating multi-modal information including depth, normal, and semantic signals derived from SMPL.

\subsection{Human-environment Affordance Generation}
Numerous studies leverage diffusion models to generate human image or video that contextually integrate with scenes or interactive objects. Some studies\cite{putting,environment-specific,addme,text2place,invi} investigate inserting or inpainting human into given scenes to achieve scene affordance. \cite{putting} applies video self-supervised training to inpaint person into masked region with correct affordances. Text2Place\cite{text2place} aims to place a person in background scenes by learning semantic masks using text guidance for localizing regions. InVi\cite{invi} achieves object insertion by first conducting image inpainting and subsequently generating frames using extended-attention mechanisms.

Several works focus on character animation with scene or object interactions. MovieCharacter\cite{moviecharacter} composites the animated character results into person-removed video sequence. AnchorCrafter\cite{anchorcrafter}, focusing on human-object interaction, first perceives HOI-appearances and injects HOI-motion to generate anchor-style product promotion videos. MIMO\cite{mimo} introduces spatial decomposed diffusion, decomposing videos into human, background and occlusion based on 3D depth and subsequently composing these elements to generate character video.

\section{Method}

In this section, we introduce \textit{Animate Anyone 2}. In \ref{sec:framework}, we first elaborate on the overall framework. In \ref{sec:scene}, we delineate the strategy for environment formulation. In \ref{sec:object}, we present the design of object injection. In \ref{sec:pose}, we provide a detailed exposition of pose modulation strategy.

\subsection{Framework}\label{sec:framework}

\noindent
\textbf{System Setting. }
The overall framework is illustrated in Fig.\ref{fig:overview}.
During training, we employ a self-supervised learning strategy. Given a reference video ${\mathit I}^{1:\mathit N}$ where $\mathit N$ denotes the number of frames, we disentangle character and environment via a formulated mask (detailed in \ref{sec:scene}), obtaining separate character sequence ${\mathit I}^{1:\mathit N}_{\mathit c}$ and environment sequence ${\mathit I}^{1:\mathit N}_{\mathit e}$. To facilitate more fidelity object interaction, we additionally extracted the sequence of objects ${\mathit I}^{1:\mathit N}_{\mathit o}$. 
Motion sequence ${\mathit I}^{1:\mathit N}_{\mathit m}$ is extracted as driving signals.
We randomly sample a character image ${\mathit I}_{\mathit c}$ from ${\mathit I}^{1:\mathit N}_{\mathit c}$ with center crop and composite it onto a random background. Given image ${\mathit I}_{\mathit c}$, motion sequence ${\mathit I}^{1:\mathit N}_{\mathit m}$, environment sequence ${\mathit I}^{1:\mathit N}_{\mathit e}$ and object sequence ${\mathit I}^{1:\mathit N}_{\mathit o}$ as inputs, our model reconstructs the reference video ${\mathit I}^{1:\mathit N}$. 
During inference, given a target character image and a driving video, our method can animate the character with consistent actions and environmental relationship corresponding to the driving video.

\noindent
\textbf{Diffusion Model. }
Our method is developed based on LDM\cite{ldm}. It employs a pretrained VAE\cite{vae,vqvae} to transform images from pixel space to latent space: $\mathbf z \mathcal = \mathcal E$($\mathbf x$). During training, random Gaussian noise $\epsilon$ is progressively added to image latents ${\mathbf z}_{t}$ at different timesteps,
The training objective can be formulated as follows:

\begin{equation}
\label{eq1}
    {\mathbf L} = {\mathbb E}_{{\mathbf z}_{t},c,{\epsilon},t}({||{\epsilon}-{{\epsilon}_{\theta}}({\mathbf z}_{t},c,t)||}^{2}_{2})
\end{equation}

\noindent
where ${\epsilon}_{\theta}$ represents the function of DenoisingNet. $\mathnormal c$ represents conditional inputs. During inference, noise latents are iteratively denoised\cite{denoising,ddim} and reconstructed into images through the decoder of VAE: ${\mathbf x}_{recon} \mathcal = \mathcal D$($\mathbf z$). 
The network design of DenoisingNet is derived from Stable Diffusion\cite{ldm}, inheriting its pretrained weights. We extend the original 2D UNet architecture to 3D UNet, incorporating the temporal layer design from AnimateDiff\cite{animatediff}.

\noindent
\textbf{Conditional Generation. }
We adopt the ReferenceNet architecture from \cite{aa} to extract appearance features of the character image ${\mathit I}_{\mathit c}$. 
In our framework, we simplify the computational complexity by merging these features exclusively in the midblock and upblock of the DenoisingNet decoder via spatial attention\cite{attention}.
Besides, three conditional embeddings are extracted from the souce video: environment sequence ${\mathit I}^{1:\mathit N}_{\mathit e}$, motion sequence ${\mathit I}^{1:\mathit N}_{\mathit m}$, and object sequence ${\mathit I}^{1:\mathit N}_{\mathit o}$. For environment sequence ${\mathit I}^{1:\mathit N}_{\mathit e}$, we employ VAE encoder to encode the embedding and subsequently merge it with noise latents. For motion sequence ${\mathit I}^{1:\mathit N}_{\mathit m}$, we design pose modulation strategy (elaborated in \ref{sec:pose}) and the motion information is also merged into the noise latents. For object sequence ${\mathit I}^{1:\mathit N}_{\mathit o}$, after encoding via VAE encoder, we develop an object guider to extract multi-scale features and inject them into the DenoisingNet through spatial blending, which will be detailed in \ref{sec:object}.

\subsection{Environment Formulation}\label{sec:scene}

\noindent
\textbf{Motivation. }
In our framework, the environment is formulated as a region excluding characters. During training, the model generates characters to populate these regions while maintaining coherence with the environmental context. 
The boundary relationship between characters and the environment is crucial. Appropriate boundary guidance can facilitate the model in learning character-environment integration more effectively, while preserving character shape consistency and environmental information integrity.
Some studies\cite{putting,invi} leverage bounding boxes to represent generative regions. 
However, we observe artifacts or inconsistencies with the source video when dealing with complex scenes, due to insufficient conditioning.
Conversely, directly using precise masks is also suboptimal, potentially introducing shape leakage.
Due to the self-supervised training strategy, there exists strong correlation between character outlines and mask boundaries. Consequently, the model tends to use this information as additional guidance for animating character. However, during inference, when the target character differs from the source in body shape and clothing, the model may forcibly conform to the mask boundary, resulting in integration artifacts.

\begin{figure}[!t]
\begin{center}
    \vspace{-0.3cm}
	\setlength{\fboxrule}{0pt}
	\fbox{\includegraphics[width=1\linewidth]{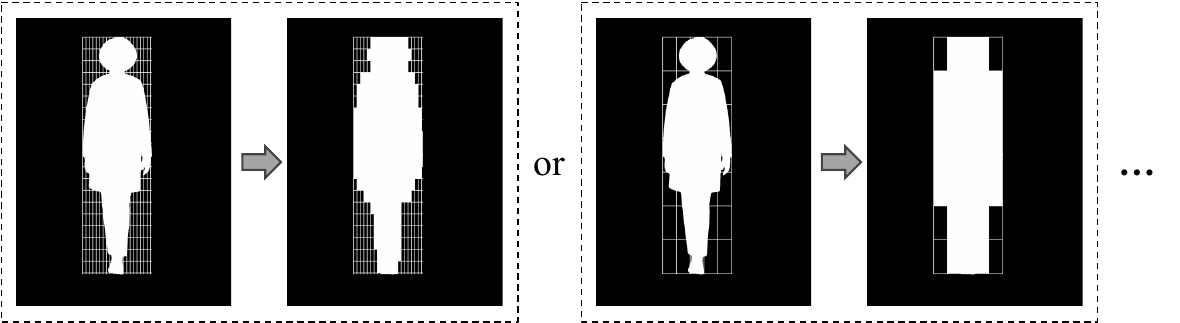}}
\end{center}
\vspace{-0.6cm}
\caption{Different coefficients for mask formulation.}
\vspace{-0.3cm}
\label{fig:mask}
\end{figure}

\noindent
\textbf{Shape-agnostic Mask. }
Therefore, we propose a shape-agnostic mask strategy for environment formulation, with the core idea of disrupting the correspondence between mask region and character outline during training. Specifically, for a character mask ${\mathit M}_{\mathit c}$ in its bounding box of size ${\mathit h} \times {\mathit w}$, we define two coefficients ${\mathit k}_{\mathit h}$ and ${\mathit k}_{\mathit w}$. 
We divided the character mask ${\mathit M}_{\mathit c}$ into ${\mathit k}_{\mathit h} \times {\mathit k}_{\mathit w}$ non-overlapping blocks, where ${\mathit k}_{\mathit h} \in (1,{\mathit h}), {\mathit k}_{\mathit w} \in (1,{\mathit w})$. 
We denote ${\mathit P}^{\mathit (k)}_{\mathit c}$ as the divided patches, where ${\mathit k}$ is the index. 
We reformulate the mask ${\mathit M}_{\mathit c}$ into a new mask ${\mathit M}_{\mathit f}$ by propagating the patch-wise maximum value:

    
\begin{equation}
\label{eq1}
{\mathit M}_{\mathit f}(i, j) = \max_{(i,j) \in  {\mathit P}^{\mathit (k)}_{\mathit c} } {\mathit P}^{\mathit (k)}_{\mathit c}(i, j)
\end{equation}


\noindent
where ${\mathit P}^{\mathit (k)}_{\mathit c}(i, j)$ represents the value at position ${\mathit (i,j)}$.
The visualized process is presented in Fig.\ref{fig:mask}. By employing this strategy, the formulated mask dynamically generate different shapes that deviate from the character boundaries, thereby compelling the network to learn context integration more effectively, unencumbered by predefined boundary constraints. During inference, we set ${\mathit k}_{\mathit h} = {\mathit h} / 10$ and ${\mathit k}_{\mathit w} = {\mathit w} / 10$.

\noindent
\textbf{Random Scale Augmentation. }
Moreover, since the formulated mask is inherently larger than the original mask, this introduces an inevitable bias that constrains the generated character to be necessarily smaller than the given mask. To mitigate this bias, we employ random scale augmentation on source videos. Specifically, we extract the character together with the interacting objects based on their masks and apply a random scaling operation. Subsequently, we recompose these scaled content back into the source video. This approach ensures that the formulated mask has a probabilistic chance of being smaller than the actual character region. During inference, the model is capable of animating the character flexibly without being constrained by the size of the mask.

\subsection{Object Injection}\label{sec:object}

\noindent
\textbf{Object Guider. }
The environment formulation strategy may potentially lead to distortion of object regions.
To enhance the preservation of object interactions, we propose to inject additional object-level features. 
Interactive objects can be extracted through two methods: 1) Leveraging VLM\cite{cogvlm,qwenvl} to obtain object localization; 2) Interactively confirming object positions via manual annotation. Then we employ SAM2\cite{sam,sam2} to extract object mask, obtaining corresponding object image and encode it into object latents via VAE encoder. A naive approach to merging object features is to directly concatenate scene and object features before feeding them into the network. However, due to the intricate relationship between characters and objects, such method struggles to handle complex human-object interactions, often falling short in capturing both human and object details. 
Thus we design an object guider to extract object-level features. 
Unlike character features that require complex modeling, objects inherently preserve visual characteristics from the source video. 
Thus we implement object guider using a lightweight fully convolutional architecture. 
specifically, object latents are downsampled four times via $3 \times 3$ Conv2D to obtain multi-scale features. The channel dimensions of these features are aligned with those in the midblock and upblock of the DenoisingNet,  facilitating subsequent feature fusion.

\noindent
\textbf{Spatial Blending. }
To recover the spatial relationships of human-object interaction, we employ spatial blending to inject features extracted by object guider into the DenoisingNet. Specifically, during the denoising process, spatial blending layer is performed after spatial attention layer. For noise latents ${\mathit z}_{\mathit noise}$ and object latents ${\mathit z}_{\mathit object}$, we concatenate their features and compute the alpha weight ${\mathit \alpha}$ through a Conv2D-Sigmoid layer. The spatial blending process can be mathematically formulated as follows:

\begin{equation}
\label{eq1}
    {\mathit \alpha} = {\mathit F}(cat({\mathit z}_{\mathit noise},{\mathit z}_{\mathit object}))
\end{equation}
\begin{equation}
\label{eq2}
    {\mathit z}_{\mathit blend} = {\mathit \alpha} \cdot {\mathit z}_{\mathit object} + (1 - {\mathit \alpha}) \cdot {\mathit z}_{\mathit noise}
\end{equation}

\noindent
where ${\mathit F}$ denotes the Conv2D-Sigmoid layer, which is initialized through zero convolution. ${\mathit z}_{\mathit blend}$ denotes the new noise latents after spatial blending.  In each stage of the DenoisingNet decoder, we alternately apply spatial attention on character features and spatial blending of object features, enabling the generation of high-fidelity results with excellent details of character-object interactions.

\subsection{Pose Modulation}\label{sec:pose}

\noindent
\textbf{Motivation. }
Animate Anyone\cite{aa} employs a skeleton representation to capture character motion and utilizes pose guider for feature modeling. However, the skeleton representation lacks explicit modeling of inter-limb spatial relationships and hierarchical dependencies. Some existing methods\cite{champ,mimo} adopt 3D mesh representations like SMPL to represent human bodies, but this tends to compromise the generalizability across characters and potentially introduces shape leakage due to its dense representation.

\noindent
\textbf{Depth-wise Pose Modulation. }
We propose to retain the skeleton signals while augmenting it with structured depth to enhance the representation of inter-limb spatial relationships. We refer to this approach as depth-wise pose modulation. For motion signals, we leverage Sapien\cite{sapiens} to extract the skeleton and depth information from the source video. The depth information is structurally processed via the skeleton to mitigate potential shape leakage in raw depth maps. Specifically, we first binarize the skeleton image to obtain skeleton mask, and subsequently extract the depth results within this masked region.
Then we employ Conv2D with the same architectural design as the pose guider\cite{aa} to process the skeleton map and structured depth map. Then we merge the structured depth information into the skeleton features through a cross-attention mechanism. The key insight behind this approach is to enable each limb to incorporate spatial characteristics from other limbs, thereby facilitating a more nuanced understanding of limb interaction relationships. 
Given that pose information extracted from wild videos may contain errors, we utilize Conv3D to model temporal motion information, enhancing inter-frame connections and mitigating the impact of erroneous signals on individual frames.


\begin{figure*}[!t]
\begin{center}
	\setlength{\fboxrule}{0pt}
	\fbox{\includegraphics[width=0.99\linewidth]{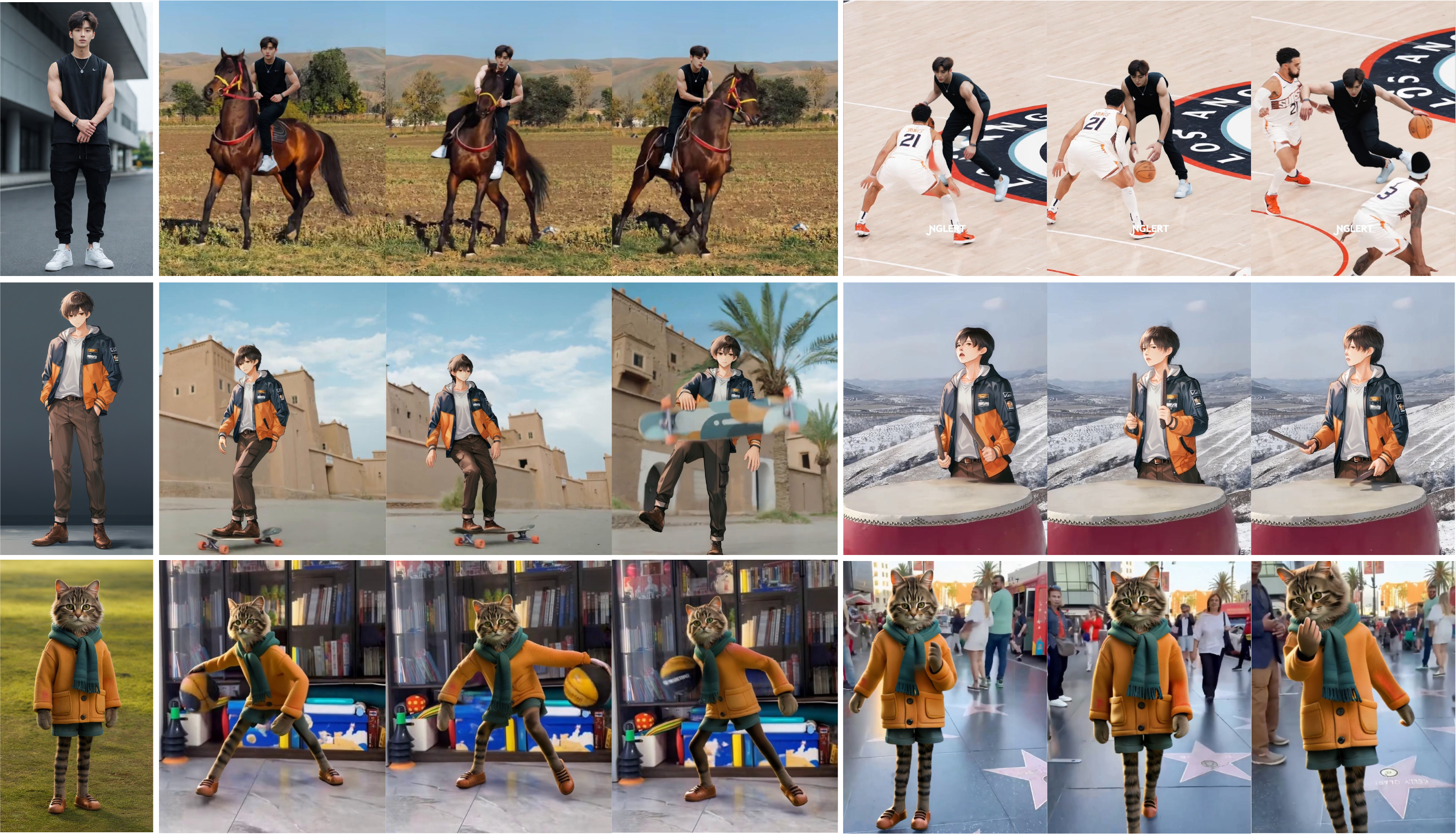}}
\end{center}
\vspace{-0.6cm}
\caption{Qualitative Results. \textit{Animate Anyone 2} achieves consistent character animation while enabling the integration and interaction between characters and their environments, thereby realizing environment affordance.}
\vspace{-0.2cm}
\label{fig:vis}
\end{figure*}

\section{Experiments}

\subsection{Implementations}

To validate the generalizability of our method across more diverse scenarios, we curated a dataset of 100,000 character videos collected from the internet, encompassing a broader range of scene types, action categories, and human-object interaction cases. Experiments are conducted on 8 NVIDIA A100 GPUs. The training involves 100k steps with batch size of 8 and the video length in a batch is 16. 
Video frames are cropped at consistent positions to ensure that the character is fully contained within the 16-frame sequence. The reference image is randomly sampled from the entire video sequence. We perform center cropping and remove the original background, compositing it with a new random background. This approach enables the model to automatically recognize characters within the image during inference without requiring additional segmentation, thereby mitigating potential accuracy limitations inherent in segmentation processes.

During long video inference, the video is segmented into multiple video clips, and inference is performed on each clip sequentially. Inspired by the motion frame technique in \cite{emo}, we utilize the final frame of the previous video clip as the temporal reference to guide the transition between clips. This strategy ensures smooth transitions between different video clips, preventing appearance texture discontinuities or blurriness.

\subsection{Qualitative Results}

Fig.~\ref{fig:vis} demonstrates that our approach not only animates diverse characters with high-fidelity performance, but also achieves remarkably seamless visual integration and interaction with their surrounding environments. This substantiates the versatility and robustness of our method, underscoring its significant potential for widespread applications.

\subsection{Comparisons}

\noindent
\textbf{Metrics. }
We follow the previous evaluation metrics for character image animation. Specifically, for single-frame quality assessment, we employ PSNR\cite{psnr}, SSIM\cite{ssim}, and LPIPS\cite{lpips}. For video fidelity, we utilize the Frechet Video Distance (FVD)\cite{fvd}.

\begin{table}
    \setlength{\tabcolsep}{4pt}
	\centering
	\resizebox{0.45\textwidth}{!}{
        \input{table/tiktok}
    }
    \vspace{-0.2cm}
	\caption{Quantitative comparison on Tiktok benchmark. * means utilizing other video data for pretraining. }
    \vspace{-0.3cm}
	\label{table:tiktok}
\end{table}

\noindent
\textbf{Evaluation on TikTok Dataset. }
We conduct experiments on the TikTok Benchmark\cite{tiktok}. In this dataset, the video backgrounds are static.
Existing character animation approaches typically synthesize target videos with both characters and backgrounds by a single reference image. To ensure a fair comparison, we adjust the configuration of our method: instead of using the ground truth background, we employ the background from the reference image as the environmental input. This modification allows all methods to generate outputs conditioned exclusively on a single reference image.
We implement two training settings of our approach: 1) trained exclusively on the Tiktok training set, and 2) first trained on our custom dataset and subsequently fine-tuned on the Tiktok training set. As shown in Tab.~\ref{table:tiktok}, when trained solely on the Tiktok training set, our method outperforms Magicanimate\cite{magicanimate} and Animate Anyone\cite{aa}. After incorporating pre-trained video data, our approach further surpasses Champ\cite{champ} and UniAnimate\cite{unianimate}, achieving state-of-the-art performance.

\begin{figure}[!t]
\begin{center}
	\setlength{\fboxrule}{0pt}
	\fbox{\includegraphics[width=0.95\linewidth]{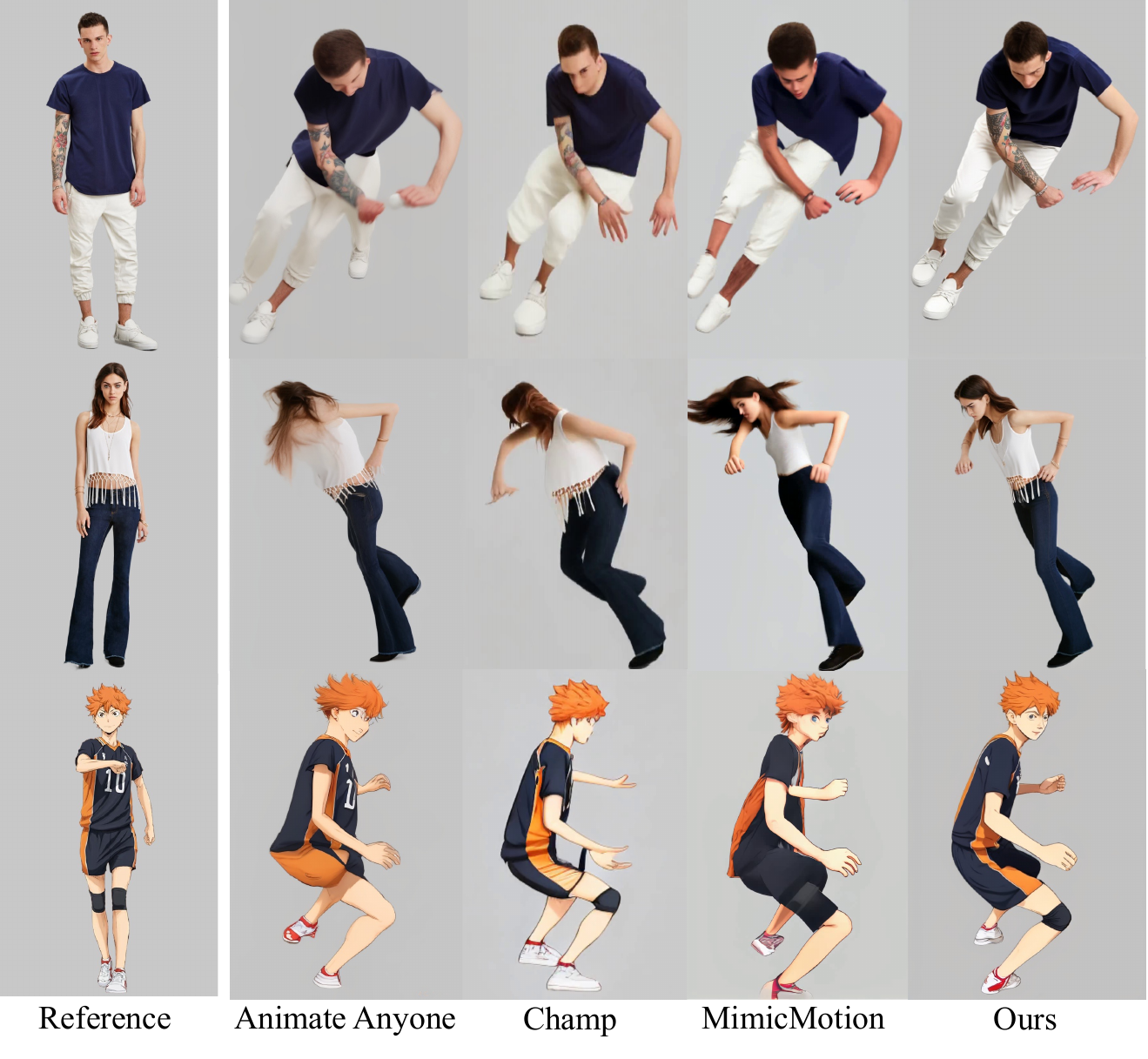}}
\end{center}
\vspace{-0.7cm}
\caption{Qualitative comparion for character animation. We normalize the background to a uniform color. }
\vspace{-0.01cm}
\label{fig:com1}
\end{figure}

\begin{table}
    \setlength{\tabcolsep}{4pt}
	\centering
        \resizebox{0.45\textwidth}{!}{
        \input{table/propose}
    }
    \vspace{-0.2cm}
	\caption{Quantitative comparison on our dataset. Our approach demonstrates superior performance across generalized scenarios.}
    \vspace{-0.2cm}
	\label{table:propose}
\end{table}

\noindent
\textbf{Evaluation on Proposed Dataset. }
Due to the limitations of existing benchmarks\cite{dwnet,tiktok,mraa} that exhibit domain proximity, these datasets cannot effectively evaluate the generalizability of models across diverse scenarios. Following ~\cite{champ}, we establish a testset comprising 100 character videos from real-world scenarios to conduct additional evaluation. Since other methods cannot generate dynamic environment, we standardize the background of input images to a uniform color, thus isolating the impact of environment variations on the evaluation. 
For fair comparison, we finetune these methods on our custom training dataset. 
The quantitative comparison is shown in Tab.~\ref{table:propose}. Qualitative comparison is shown in Fig.\ref{fig:com1}.
Our results significantly outperform alternative approaches, which can be attributed to two key factors: (1) our proposed motion modeling demonstrates robust generalization across diverse motion patterns, and (2) our decoupled environment and character generation strategy enables the model to focus more precisely on character dynamics, mitigating interference from environment variations.

\begin{figure}[!t]
\begin{center}
	\setlength{\fboxrule}{0pt}
	\fbox{\includegraphics[width=0.99\linewidth]{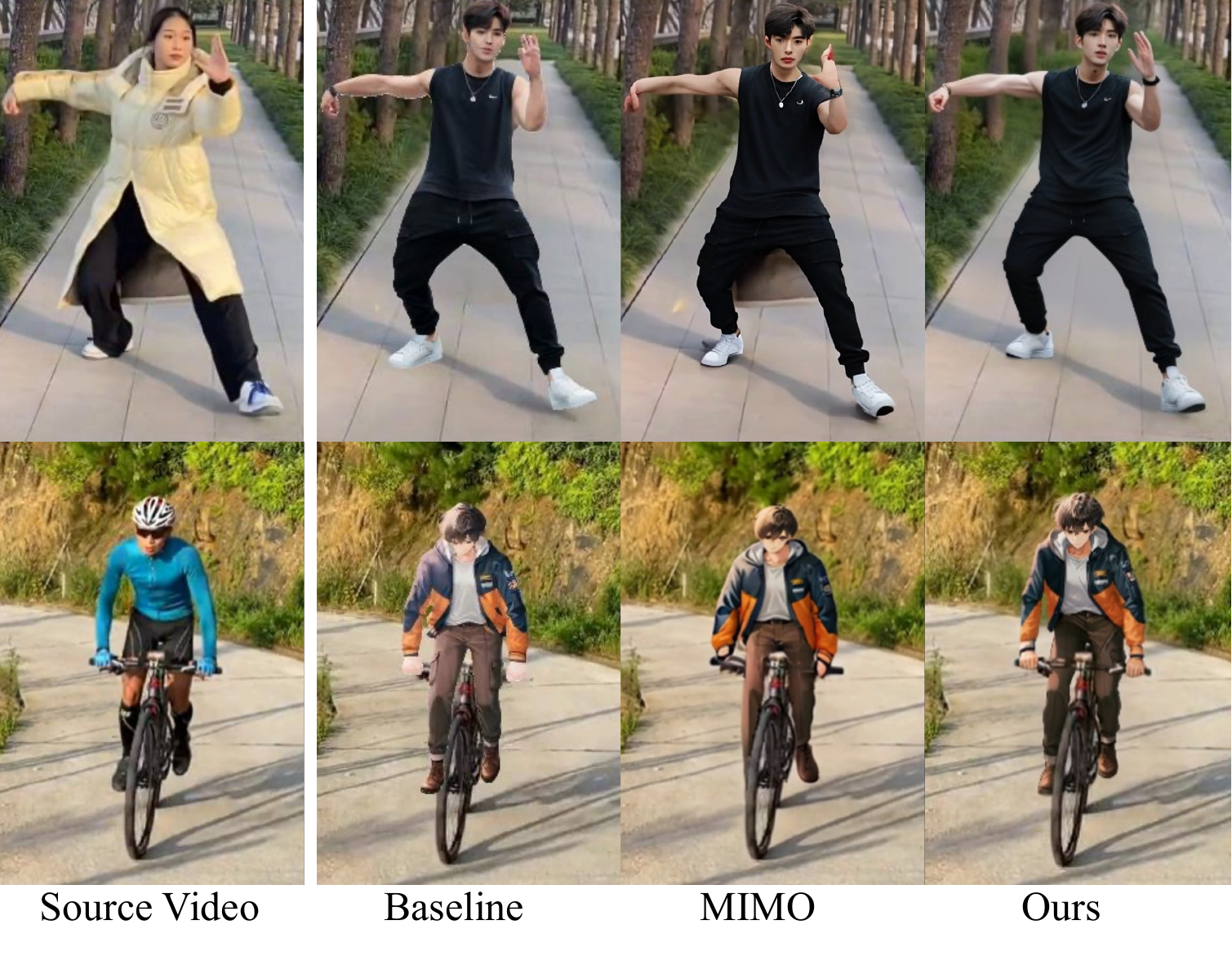}}
\end{center}
\vspace{-0.7cm}
\caption{Qualitative comparion. Our method demonstrates superior environment integration and object interaction.}
\vspace{-0.1cm}
\label{fig:mimo}
\end{figure}

\noindent
\textbf{Evaluation for character-environment affordance. }
We further evaluate the performance of character-environment affordance on our proposed dataset. 
We construct a baseline algorithm by directly compositing character animation results onto the original video background, creating a pseudo character-environment integration, similar to MovieCharacter \cite{moviecharacter}. we leverage ProPainter\cite{propainter} to inpaint the character region. 
Quantitative evaluation is presented in Tab.~\ref{table:base}. 
We conduct qualitative comparison illustrated in Fig.~\ref{fig:mimo}. Our approach demonstrates superior performance in terms of enhanced character-environment integration.
We also compare our method with MIMO\cite{mimo}, which is the most relevant method to our task setting. Due to the absence of public source code, we conduct a qualitative comparison focused on character-environment integration performance. The result of MIMO are obtained from its official ModelScope link$^*$. As illustrated in Fig.~\ref{fig:mimo}. From the first group of the visualization, it can be observed that due to MIMO's reliance on additional pre-processing algorithms for background inpainting, it tends to leave noticeable preprocessing artifacts and establish erroneous relationships between the background and the animated characters. In contrast, our proposed approach effectively mitigates these issues, enabling superior scene and character integration. The second group further illustrates MIMO's limitations in handling relatively complex human-object interaction scenarios, whereas our method demonstrates enhanced robustness in intricate scenes.

\renewcommand{\thefootnote}{}
\footnotetext{$^*$https://modelscope.cn/studios/iic/MIMO}

\begin{table}
    \setlength{\tabcolsep}{4pt}
	\centering
        \resizebox{0.36\textwidth}{!}{
        \input{table/baseline}
    }
    \vspace{-0.2cm}
	\caption{Quantitative comparison with baseline on our dataset. Baseline refers to the pseudo character-environment integration.}
    \vspace{-0.3cm}
	\label{table:base}
\end{table}

\subsection{Ablation Study}

\begin{figure}[!t]
\begin{center}
	\setlength{\fboxrule}{0pt}
	\fbox{\includegraphics[width=1\linewidth]{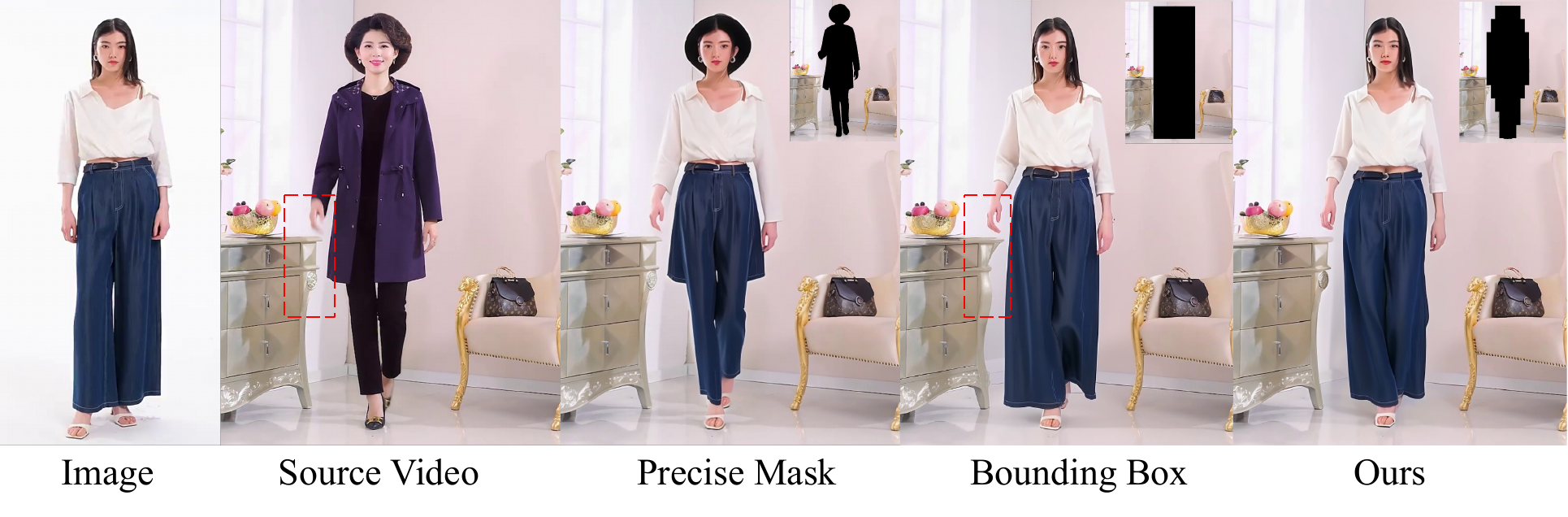}}
\end{center}
\vspace{-0.7cm}
\caption{Ablation study of environment formulation.}
\vspace{-0.1cm}
\label{fig:aba1}
\end{figure}

\noindent
\textbf{Environment Formulation. }
To demonstrate the effectiveness of our proposed environment formulation strategy, we explore alternative designs, including: 1) utilizing precise character masks from the source video, and 2) employing bounding box regions. Qualitative results are shown in Fig.~\ref{fig:aba1}. Using accurate masks can constrain the animated character's shape within the predefined mask boundaries, potentially causing appearance deformation and inconsistency. Conversely, adopting bounding box regions may introduce scene context distortions and fusion artifacts in the proximity of character. 
Our method demonstrates superior capability in learning flexible character generation and environmental completion, achieving both character consistency and seamless character-scene integration.

\begin{figure}[!t]
\begin{center}
	\setlength{\fboxrule}{0pt}
	\fbox{\includegraphics[width=0.8\linewidth]{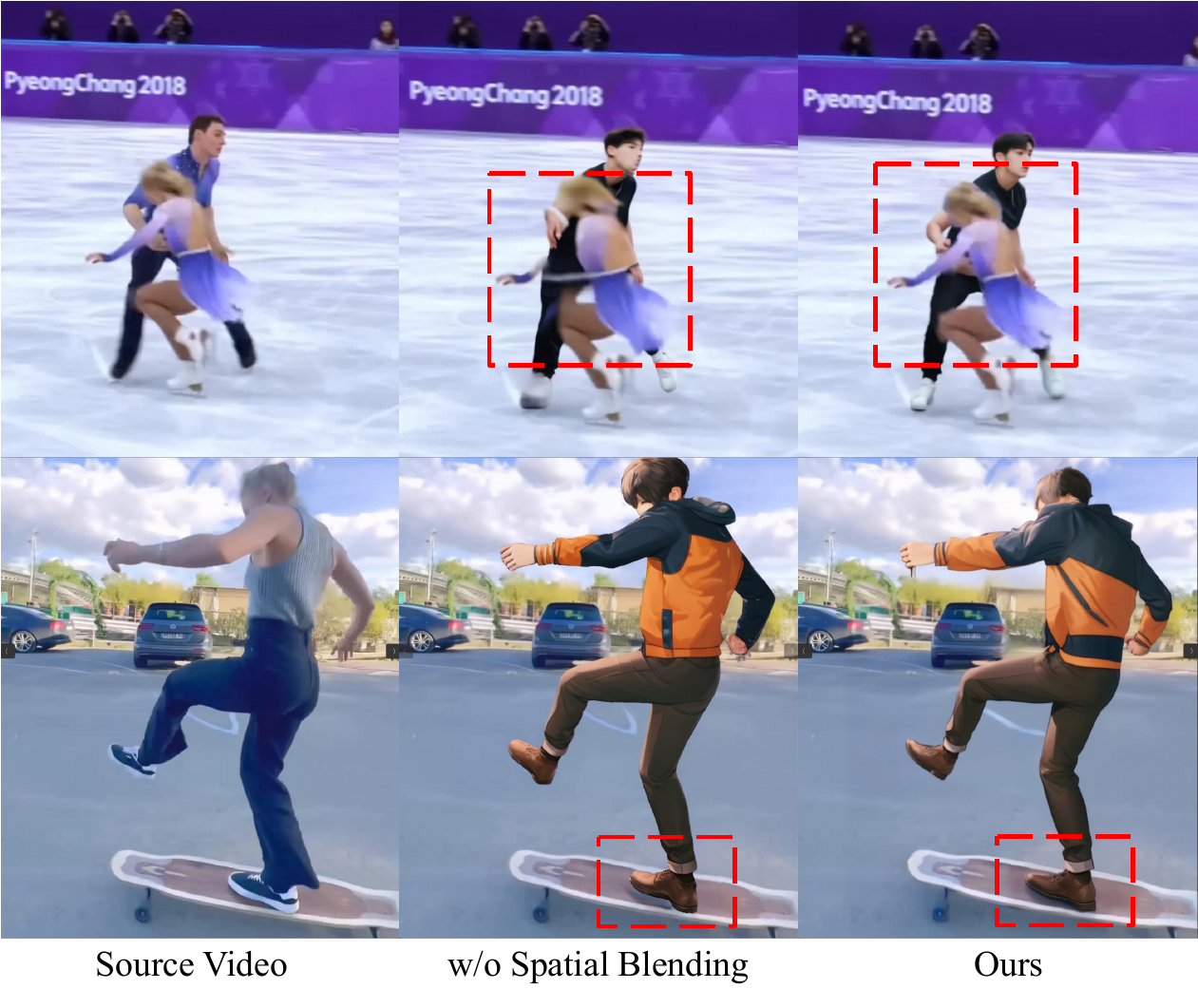}}
\end{center}
\vspace{-0.7cm}
\caption{Qualitative ablation of object modeling method.}
\vspace{-0.2cm}
\label{fig:aba2}
\end{figure}

\noindent
\textbf{Object Modeling. }
We conduct a comparison of different object modeling approaches: directly merging object features with noise latents without employing spatial blending. Quantitative result is shown in Tab~\ref{table:ablation}. We further demonstrate the visualization results. As shown in Fig.~\ref{fig:aba2}, 
in complex interaction scenarios, it fails to comprehensively preserve the intrinsic features of interactive objects, resulting in local distortions and consequently misinterpreting their interaction relationships.
The second comparison reveals that the interactions between characters and objects exhibit an artificial stitching effect, which consequently compromises the naturalness of their interactive relationships.

\begin{figure}[!t]
\begin{center}
	\setlength{\fboxrule}{0pt}
	\fbox{\includegraphics[width=0.8\linewidth]{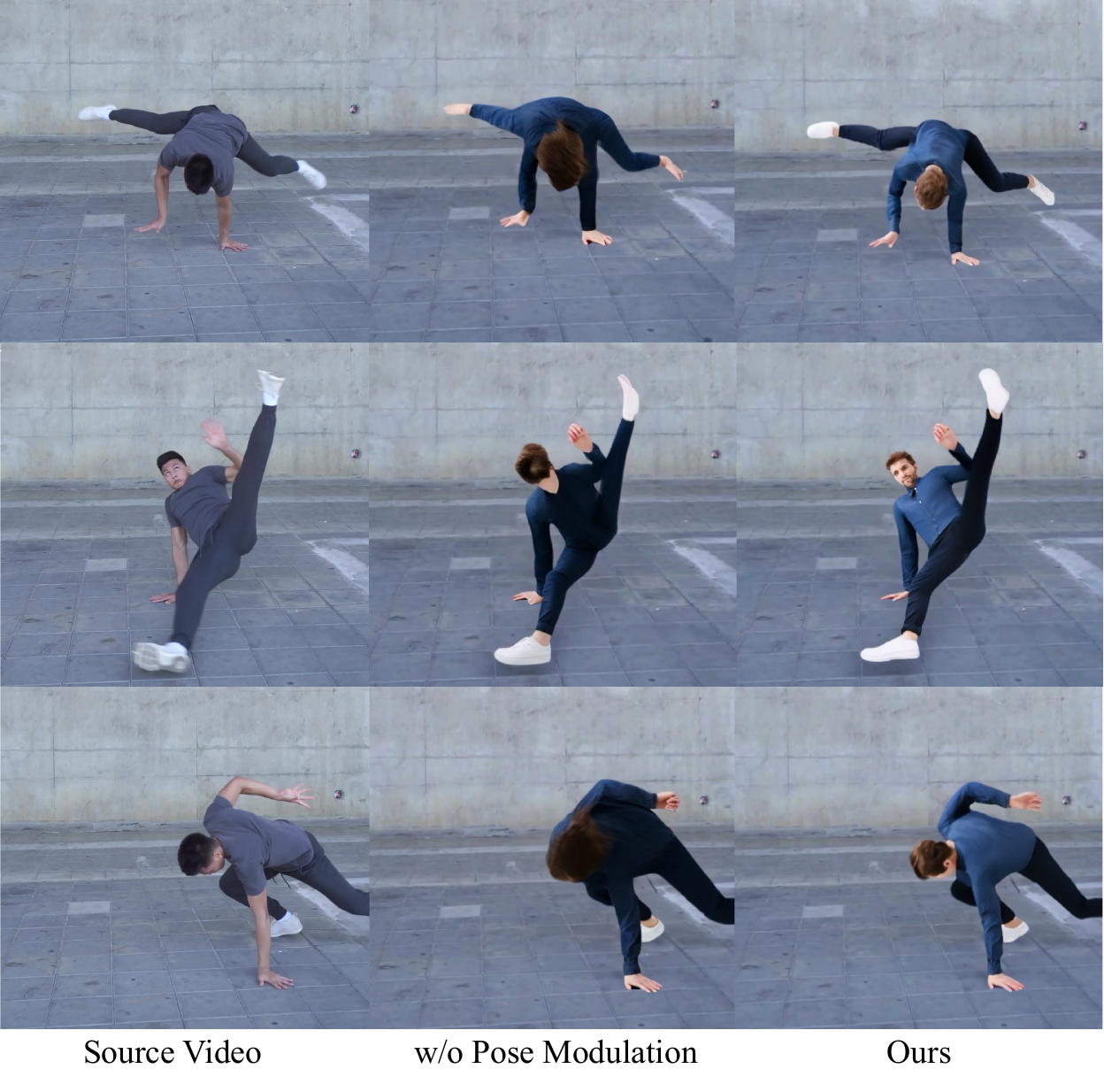}}
\end{center}
\vspace{-0.7cm}
\caption{Qualitative ablation of pose modulation.}
\vspace{-0.01cm}
\label{fig:aba3}
\end{figure}

\noindent
\textbf{Pose Modulation. }
We evaluate the effectiveness of our proposed pose modulation strategy. Quantitative result is presented in Tab~\ref{table:ablation}. Qualitative result is shown in Fig~\ref{fig:aba3}. Without employing the pose modulation method, character limb relationships may suffer from misalignment and spatial inconsistencies. Consequently, the model's capability to generate accurate and plausible character poses becomes severely constrained. In contrast, our proposed approach, by incorporating depth-aware information, can more effectively learn and capture the complex spatial relationships between limbs, enabling robust performance across diverse and challenging motion scenarios.

\begin{table}
    \setlength{\tabcolsep}{4pt}
	\centering
        \resizebox{0.45\textwidth}{!}{
        \input{table/ablation}
    }
    \vspace{-0.2cm}
	\caption{Quantitative ablation study.}
    \vspace{-0.1cm}
	\label{table:ablation}
\end{table}



\section{Discussion and Conclusion}
\noindent
\textbf{Limitations. }
Our approach may introduce visual artifacts when dealing with complex hand-object interactions that occupy a relatively small pixel region. In intricate human-object interactions, deformation artifacts may emerge when source and target characters exhibit substantial shape discrepancies. The performance of object interaction is also influenced by SAM's segmentation capabilities.

\noindent
\textbf{Potential Impact. }
The proposed method may be used to produce fake videos of individuals, which can be detected using some anti-spoofing techniques\cite{anti_color,anti_deep,anti_search,demamba,dormant}. 

\noindent
\textbf{Conclusion. }
In this paper, we introduce \textit{Animate Anyone 2}, a novel framework that enables character animation to exhibit environment affordance. We extract environmental information from driving videos, enabling the animated character to preserve its original environment. We propose a novel environment formulation and object injection strategy, facilitating seamless character-environment integration. Moreover, we propose pose modulation that empowers the model to robustly handle diverse motion patterns. 
Experimental results demonstrate that \textit{Animate Anyone 2} achieves high-fidelity generation performance.

{
    \small
    \bibliographystyle{ieeenat_fullname}
    \bibliography{main}
}


\end{document}

%% file: table/tiktok.tex
\begin{tabular}{@{}ccccc@{}}
\toprule
Method         & SSIM $\uparrow$                      & PSNR $\uparrow$                     & LPIPS $\downarrow$                    & FVD  $\downarrow$                      \\ \midrule
MRAA \cite{mraa}          & 0.672           & 29.39        & 0.672           & 284.82                     \\
DisCo \cite{disco}         &  0.668        &     29.03        & 0.292         & 292.80                     \\
MagicAnimate \cite{magicanimate}  & 0.714         & 29.16           & 0.239        & 179.07                     \\
Animate Anyone \cite{aa} & 0.718          & 29.56        & 0.285           & 171.90                      \\
Champ* \cite{champ}         & 0.802          &  29.91         & 0.234         & 160.82 \\
UniAnimate* \cite{unianimate} & 0.811           & 30.77            & 0.231        & 148.06   \\
Ours  & 0.778           & 29.82            & 0.248        & 158.97   \\
Ours*           & \textbf{0.812}            & \textbf{30.82}            & \textbf{0.223}            & \textbf{144.65}            \\ \bottomrule
\end{tabular}

%% file: table/propose.tex
\begin{tabular}{@{}ccccc@{}}
\toprule
Method         & SSIM $\uparrow$                      & PSNR $\uparrow$                     & LPIPS $\downarrow$                    & FVD  $\downarrow$                      \\ \midrule
Animate Anyone\cite{aa} & 0.761          & 28.41        & 0.324           & 228.53                      \\
Champ\cite{champ}          & 0.771          &  28.69         & 0.294         & 205.79 \\
MimicMotion\cite{mimicmotion}  & 0.767           & 28.52            & 0.307        & 212.48   \\
Ours           & \textbf{0.809}            & \textbf{29.24}            & \textbf{0.259}            & \textbf{172.54}            \\ \bottomrule
\end{tabular}

%% file: table/baseline.tex
\begin{tabular}{@{}ccccc@{}}
\toprule
Method         & SSIM $\uparrow$                      & PSNR $\uparrow$                     & LPIPS $\downarrow$                    & FVD  $\downarrow$                      \\ \midrule
Baseline  & 0.785           & 28.71            & 0.291        & 195.45   \\
Ours           & \textbf{0.794}            & \textbf{28.83}            & \textbf{0.276}            & \textbf{186.17}            \\ \bottomrule
\end{tabular}

%% file: table/ablation.tex
\begin{tabular}{@{}ccccc@{}}
\toprule
Method & SSIM $\uparrow$ & PSNR $\uparrow$& LPIPS $\downarrow$& FVD  $\downarrow$                      \\ \midrule
w/o Spatial Blending  & 0.789           & 28.74            & 0.283        & 191.23   \\
w/o Pose Modulation  & 0.769           & 28.56            & 0.301        & 211.15   \\
Ours           & \textbf{0.794}            & \textbf{28.83}            & \textbf{0.276}            & \textbf{186.17}            \\ \bottomrule
\end{tabular}